%% file: main.tex
\documentclass[11pt]{article}

\usepackage{booktabs}
\usepackage{luca}
\usepackage{graphicx}
\usepackage{amsfonts,amsmath,amssymb}
\usepackage{url}
\usepackage{hyperref}
\usepackage{comment} 
\usepackage{float} 
\usepackage{tablefootnote} 
\usepackage{xcolor}
\usepackage{geometry} 
\usepackage{rbf}
\usepackage{subfig}

\begin{document}
\title{Neural Networks with Structural Resistance
to Adversarial Attacks}
\author{Luca de Alfaro \\[1ex]
\normalsize Computer Science and Engineering Department\\
\normalsize University of California, Santa Cruz\\
\normalsize luca@ucsc.edu}
\date{\today \\
  \normalsize Technical Report UCSC-SOE-18-13\\
  \normalsize School of Engineering, University of California, Santa Cruz}
\maketitle

\begin{abstract}

In adversarial attacks to machine-learning classifiers, small perturbations are added to input that is correctly classified.
The perturbations yield adversarial examples, which are virtually indistinguishable from the unperturbed input, and yet are misclassified. 
In standard neural networks used for deep learning, attackers can craft adversarial examples from most input to cause a misclassification of their choice. 

We introduce a new type of network units, called \rbfinf\ units, whose non-linear structure makes them inherently resistant to adversarial attacks. 
On permutation-invariant MNIST, in absence of adversarial attacks, networks using RBFI units match the performance of networks using sigmoid units, and are slightly below the accuracy of networks with ReLU units. 
When subjected to adversarial attacks, networks with RBFI units retain accuracies above 90\% for attacks that degrade the accuracy of networks with ReLU or sigmoid units to below 2\%. 
RBFI networks trained with regular input are superior in their resistance to adversarial attacks even to ReLU and sigmoid networks trained with the help of adversarial examples.

The non-linear structure of RBFI units makes them difficult to train using standard gradient descent. 
We show that networks of RBFI units can be efficiently trained to high accuracies using {\em pseudogradients,} computed using functions especially crafted to facilitate learning instead of their true derivatives.
We show that the use of pseudogradients makes training deep RBFI networks practical, and we compare several structural alternatives of RBFI networks for their accuracy.
\end{abstract}

\input{intro}

\input{related}
\input{rbfnets}
\input{adversaries}

\input{experiments}

\input{conclusions}

\input{main.bbl}

\end{document}

%% file: intro.tex
\section{Introduction}

Machine learning via deep neural networks has been remarkably successful in a wide range of applications, from speech recognition to image classification and language processing. 
While very successful, deep neural networks are affected by adversarial examples: small, especially crafter modifications of correctly classified input that are misclassified \cite{SzegedyIntriguingpropertiesneural2013}. 
The trouble with adversarial examples is that the modifications to regular input are so small as to be difficult or impossible to detect for a human: this has been shown both in the case of images \cite{SzegedyIntriguingpropertiesneural2013,NguyenDeepneuralnetworks2015a} and sounds \cite{KurakinAdversarialexamplesphysical2016,CarliniAudioadversarialexamples2018}. 
Further, the adversarial examples are in some measure transferable from one neural network to another \cite{GoodfellowExplainingharnessingadversarial2014,NguyenDeepneuralnetworks2015a,Papernotlimitationsdeeplearning2016,Tramerspacetransferableadversarial2017}, so they can be crafted even without precise knowledge of the weights of the target neural network. 
Together, these two properties make it possible to craft malicious and undetectable attacks for a wide range of applications that rely on deep neural networks. 
At a fundamental level, it is hard to provide guarantees about the behavior of a deep neural network, when every correctly classified input is tightly encircled by very similar, yet misclassified, inputs. 

Thus far, the approach for obtaining neural networks that are more resistant to adversarial attacks has been to feed to the networks, as training data, an appropriate mix of the original training data, and adversarial examples \cite{GoodfellowExplainingharnessingadversarial2014,MadryDeepLearningModels2017}.
In training neural networks using adversarial examples, if the examples are generated via efficient heuristics such as the {\em fast gradient sign method,} the networks learn to associate the specific adversarial examples to the original input from which they were derived, in a phenomenon known as {\em label leaking\/} \cite{KurakinAdversarialmachinelearning2016,MadryDeepLearningModels2017,TramerEnsembleadversarialtraining2017}. 
This does not result in increased resistance to general adversarial attacks \cite{MadryDeepLearningModels2017,Carlinievaluatingrobustnessneural2017}. 
If the adversarial examples used in training are generated via more general optimization techniques, as in \cite{MadryDeepLearningModels2017}, networks with markedly increased resistance to adversarial attacks can be obtained, at the price of a more complex and computationally expensive training regime, and an increase in required network capacity.

We pursue here a different approach, proposing the use of neural network types that are, due to their structure, inherently impervious to adversarial attacks, even when trained on standard input only.
In \cite{GoodfellowExplainingharnessingadversarial2014}, the authors connect the presence of adversarial examples to the (local) linearity of neural networks. 
In a purely linear form $\sum_{i=1}^n x_i w_i$, we can perturb each $x_i$ by $\epsilon$, taking $x_i + \epsilon$ if $w_i > 0$, and $x_i - \epsilon$ if $w_i < 0$. 
This causes an output perturbation of magnitude $\epsilon \sum_{i=1}^n |w_i|$, or $n \bar{w}$ for $\bar{w}$ the average modulus of $w_i$. 
When the number of inputs $n$ is large, as is typical of deep neural networks, a small input perturbation can cause a large output change. 
Of course, deep neural networks are not globally linear, but the insight of \cite{GoodfellowExplainingharnessingadversarial2014} is that they may be sufficiently locally linear to allow adversarial attacks. 
Following this insight, we develop networks composed of units that are highly non-linear. 

The networks on which we settled after much experimentation are a variant of the well known {\em radial basis functions\/} (RBFs) \cite{BroomheadRadialbasisfunctions1988,ChenOrthogonalleastsquares1991,OrrIntroductionradialbasis1996}; we call our variant RBFI units. 
RBFI units are similar to classical Gaussian RBFs, except for two differences that are crucial in obtaining both high network accuracy, and high resistance to attacks. 
First, rather than being radially symmetrical, RBFIs can scale each input component individually; in particular, they can be highly sensitive to some inputs while ignoring others. 
This gives an individual RBFI unit the ability to cover more of the input space than its symmetrical variants. 
Further, the distance of an input from the center of the Gaussian is measured not in the Euclidean, or $\thenorm_2$, norm, but in the infinity norm $\thenorm_\infty$, which is equal to the maximum of the differences of the individual components. 
This eliminates all multi-input linearity from the local behavior of a RBFI: at any point, the output depends on one input only; the $n$ in the above discussion is always $1$ for RBFIs, so to say.
The ``I'' in RBFI stands for the infinity norm. 

Using deeply nonlinear models is hardly a new idea, but the challenge has been that such models are typically very difficult to train. 
Indeed, we show that networks with RBFI units cannot be easily trained using gradient descent. 
To get around this, we show that the networks can be trained efficiently, and to high accuracy, using {\em pseudogradients.} 
A {\em pseudogradient\/} is computed just as an ordinary gradient, except that we artificially pretend that some functions have a derivative that is different from the true derivative, and especially crafted to facilitate training. 
In particular, we use pseudoderivatives for the exponential function, and for the maximum operator, that enter the definition of Gaussian RBFI units. 
Gaussians have very low derivative away from their center, which makes training difficult; our pseudoderivative artificially widens the region of detectable gradient around the Gaussian center. 
The maximum operator has non-zero derivative only for one of its inputs at a time; we adopt a pseudogradient that propagates back the gradient to all of its inputs, according to their proximity in value to the maximum input. 
Tampering with the gradient may seem unorthodox, but methods such as AdaDelta \cite{ZeilerADADELTAadaptivelearning2012}, and even gradient descent with momentum, cause training to take a trajectory that does not follow pure gradient descent. 
We simply go one step further, devising a scheme that operates at the granularity of the individual unit.

We show that with these two changes, RBFIs can be easily trained with standard random (pseudo)gradient descent methods, yielding networks that are both accurate, and resistant to attacks. 
Specifically, we consider {\em permutation invariant MNIST,} which is a version of MNIST in which the $28 \times 28$ pixel images are flattened into a one-dimensional vector of $784$ values and fed as a feature vector to neural networks \cite{GoodfellowExplainingharnessingadversarial2014}.
On this test set, we show that for nets of 512,512,512,10 units, RBFI networks match the classification accuracy of networks of sigmoid units ($(96.96 \pm 0.14)\%$ for RBFI vs.\ $(96.88 \pm 0.15)\%$ for sigmoid), and are close to the performance of network with ReLU units ($(98.62 \pm 0.08)\%$). 
When trained over standard training sets, RBFI networks retain accuracies well over 90\% for adversarial attacks that reduce the accuracy of ReLU and sigmoid networks to below $2\%$ (worse than random). 
We show that RBFI networks trained on normal input are superior to ReLU and sigmoid networks trained even with adversarial examples.  
Our experimental results can be summarized as follows:
\begin{itemize}
\item In absence of adversarial attacks, RBFI networks match the accuracy of sigmoid networks, and are slightly lower in accuracy than ReLU networks.
\item When networks are trained with regular input only, RBFI networks are markedly more resistant to adversarial attacks than sigmoid or ReLU networks.
\item In presence of adversarial attacks, RBFI networks trained on regualar input provide higher accuracy than sigmoid or ReLU networks, even when the latter are trained also on adversarial examples, and even when the adversarial examples are obtained via general projected gradient descent \cite{MadryDeepLearningModels2017}. 
\item RBFI networks can be successfully trained with pseudogradients; the training via standard gradient descent yields instead markedly inferior results.
\item Appropriate regularization helps RBFI networks gain increased resistance to adversarial attacks.
\end{itemize}
Of course, much work remains to be done, including experimenting with convolutional networks using RBFI units for images. 
However, the results seem promising, in that RBFI seem to offer a viable alternative to current adversarial training regimes for applications where resistance to adversarial attacks is important. 

To conduct our experiments, we have implemented RBFI networks on top of the PyTorch framework \cite{PaszkeAutomaticdifferentiationpytorch2017}; all the code used in their implementation and in the experiments performed in this paper is available at \url{\codeurl}.

%% file: related.tex
\section{Related Work}

Adversarial examples were first noticed in \cite{SzegedyIntriguingpropertiesneural2013}, where they were generated via the solution of general optimization problems. 
In \cite{GoodfellowExplainingharnessingadversarial2014}, a connection was established between linearity and adversarial attacks. 
A fully linear form $\sum_{i=1}^n x_i w_i$ can be perturbed by using $x_i + \epsilon \, \sign(w_i)$, generating an output change of magnitude $\epsilon \cdot \sum_{i=1}^n |w_i|$. 
In analogy, \cite{GoodfellowExplainingharnessingadversarial2014} introduced the {\em fast gradient sign method} (FGSM) method of creating adversarial perturbations, by taking $x_i + \epsilon \cdot \sign(\grad_i \call)$, where $\grad_i \call$ is the loss gradient with respect to input $i$. 
The work also showed how adversarial examples are often transferable across networks, and it asked the question of whether it would be possible to construct non-linear structures, perhaps inspired by RBFs, that are less linear and are more robust to adversarial attacks. 
This entire paper is essentially a long answer to the conjectures and suggestions expressed in \cite{GoodfellowExplainingharnessingadversarial2014}. 

It was later discovered that training on adversarial examples generated via FGSM does not confer strong resistance to attacks, as the network learns to associate the specific examples generated by FGSM to the original training examples in a phenomenon known as {\em label leaking\/} \cite{KurakinAdversarialmachinelearning2016,MadryDeepLearningModels2017,TramerEnsembleadversarialtraining2017}.
The FGSM method for generating adversarial examples was extended to an iterative method, I-FGSM, in \cite{KurakinAdversarialexamplesphysical2016}.
In \cite{TramerEnsembleadversarialtraining2017}, it is shown that using small random perturbations before applying FSGM enhances the robustness of the resulting network. 
The network trained in \cite{TramerEnsembleadversarialtraining2017} using I-FSGM and ensemble method won the first round of the NIPS 2017 competition on defenses with respect to adversarial attacks. 

Carlini and Wagner in a series of papers show that training regimes based on generating adversarial examples via simple heuristics, or combinations of these, in general fail to convey true resistance to attacks \cite{CarliniAdversarialexamplesare2017,Carlinievaluatingrobustnessneural2017}.
They further advocate measuring the resistance to attacks with respect to attacks found via more general optimization processes. 
In particular, FGSM and I-FGSM rely on the local gradient, and training techniques that break the association between the local gradient and the location of adversarial examples makes networks harder to attack via FGSM and I-FGSM, without making the networks harder to attack via general optimization techniques. 
In this paper, we follow this suggestion by using a general optimization method, projected gradient descent (PGD), to generate adversarial attacks and evaluate network robustness. 
\cite{CarliniDefensivedistillationnot2016,Carlinievaluatingrobustnessneural2017} also shows that the technique of {\em defensive distillation,} which consists in appropriately training a neural network on the output of another \cite{PapernotDistillationdefenseadversarial2016}, protects the networks from FGSM and I-FGSM attacks, but does not improve network resistance in the face of general adversarial attacks. 

In \cite{MadryDeepLearningModels2017} it is shown that by training neural networks on adversarial examples generated via PGD, it is possible to obtain networks that are genuinely more resistant to adversarial examples.
The price to pay is a more computationally intensive training, and an increase in the network capacity required. 
We provide an alternative way of reaching such resistance, one that does not rely on a new training regime. 

%% file: rbfnets.tex
\section{RBFI Units and RBFI Networks}

Consider a classifier $f$ that given an input feature vector $\vecx \in \reals^n$ generates a classification $f(x) \in \set{1, \ldots, K}$, where $n$ is the dimension of the input, and $K$ is the number of classes. 
An {\em adversarial attack\/} for a correctly-classified input $\vecx$ consists in an input $\vecx'$ close to $\vecx$ in a metric of choice, and such that $f(\vecx) \neq f(\vecx')$.
We choose in this paper the infinity norm, following \cite{GoodfellowExplainingharnessingadversarial2014,MadryDeepLearningModels2017}, so that an $\eta$-adversarial attack is one in which $\norm{\infty}{\vecx - \vecx'} \leq \eta$.

\subsection{\rbfinf\ Units}

In \cite{GoodfellowExplainingharnessingadversarial2014}, the adversarial attacks are linked to the linearity of the models. 
For this reason, we seek to use units that do not exhibit a marked linear behavior, and specifically, units which yield small output variations for small variations of their inputs measured in infinity norm 

A linear form $g(\vecx) = \sum_i x_i w_i$ represents the norm-2 distance of the input vector $x$ to a hyperplane perpendicular to vector $\vecw$, scaled by $|\vecw|$.  
In our quest for robustness, we may seek to replace this norm-2 distance with an infinity-norm distance. 
However, it turns out that the infinity-norm distance of a point from a plane is not a generally useful concept: it is preferable to consider the infinity-norm distance between points.

Hence, we define our units as variants of the classical Gaussian {\em radial basis functions\/} \cite{BroomheadRadialbasisfunctions1988a,OrrIntroductionradialbasis1996}. 
We call our variant \rbfinf, to underline the fact that they are built using infinity norm.
An \rbfinf\ unit $\neuron(\vecu, \vecw)$ for an input in $\reals^n$ is parameterized by two vectors of weights $\vecu = \tuple{u_1, \ldots, u_n}$ and $\vecw = \tuple{w_1, \ldots, w_n}$
Given an input $\vecx \in \reals^n$, the unit produces output 
\begin{equation} \label{eq-rbf}
  \neuron_\gamma(\vecu, \vecw)(\vecx) = \exp\left(- \norm{\gamma}{\vecu \hadamard (\vecx - \vecw)}^2\right) \eqpun ,
\end{equation}
where $\hadamard$ is the Hadamard, or element-wise, product, and where $\norm{\gamma}{\cdot}$ indicates the $\gamma$-norm.
In (\ref{eq-rbf}), the vector $\vecw$ is a point from which the distance to $\vecx$ is measured in $\gamma$-norm, and the vector $\vecu$ provides scaling factors for each coordinate.
Without loss of expressiveness, we require the scaling factors to be non-negative, that is, $u_i \geq 0$ for all $1 \leq i \leq n$. 
The scaling factors provide the flexibility of disregarding some inputs $x_i$, by having $u_i \approx 0$, while emphasizing the influence of other inputs. 
As we are interested here in robustness with respect to the infinity norm, our \rbfinf\ units are obtained by taking $\gamma = \infty$, in which case (\ref{eq-rbf}) can be written as:
\begin{equation} \label{eq-rbfinf}
  \neuron_\infty(\vecu, \vecw)(\vecx) = \exp\left(- \max_{1 \leq i \leq n} \bigl(u_i (x_i - w_i)\bigr)^2 \right) \eqpun . 
\end{equation}
The output of a \rbfinf\ unit is close to $1$ only when $\vecx$ is close to $\vecw$ in the coordinates that have large scaling factors. 
Thus, the unit is reminiscent of an And gate, with normal or complemented inputs, which outputs $1$ only for one value of its inputs. 
Logic circuits are composed both of And and of Or gates. 
Thus, we introduce an Or \rbfinf\ unit by: 
\begin{equation} \label{eq-rbfinf-or}
  \neuron^{-}_\infty(\vecu, \vecw)(\vecx) = 1 - \exp\left(- \max_{1 \leq i \leq n} \bigl(u_i (x_i - w_i)\bigr)^2 \right) \eqpun . 
\end{equation}
We construct neural networks out of \rbfinf\ units using three types of layers: 
\begin{itemize}

\item {\em And layer:} all units in the layer are And units, defined by (\ref{eq-rbfinf}).
\item {\em Or layer:} all units in the layer are Or units, defined by (\ref{eq-rbfinf-or}).
\item {\em Mixed layer:} the units of the layer are a mix of Or and And \rbfinf\ units.  When the network is initialized, the And or Or-ness of each unit is chosen at random, and henceforth it is kept fixed during training and classification.
\end{itemize}
We will provide a comparison of the performance of networks using different types of layers. 
Obviously, layers using \rbfinf\ units can be mixed with layers using other types of units. 

\subsection{Sensitivity Bounds for Adversarial Attacks}
\label{subsec-sensitivity}

To form an intuitive idea of why networks with \rbfinf\ units might resist adversarial attacks, it is useful to compute the sensitivity of individual units to such attacks. 
For $x \in \reals^n$ and $\epsilon > 0$, let $B_\epsilon(x) = \set{x' \mid \norm{\infty}{x - x'} \leq \epsilon}$ be the set of inputs within distance $\epsilon$ from $x$ in infinity norm.
Given a function $f: \reals^n \mapsto \reals$, we call its {\em sensitivity to adversarial attacks\/} the quantity:
\begin{equation} \label{eq-sensitivity-def}
   s = \sup_{x \in \reals^n} \, \limsup_{\epsilon \goto 0} 
     \frac{\sup_{x' \in B_\epsilon(x)} | f(x) - f(x')|}{\epsilon} \eqpun .
\end{equation}
The sensitivity (\ref{eq-sensitivity-def}) represents the maximum change in output we can obtain via an input change within $\epsilon$ in infinity norm, as a multiple of $\epsilon$ itself. 
For a single ReLU unit with weight vector $\vecw$, the sensitivity is given by 
\begin{equation} \label{eq-sensitivity-relu}
  s = \sum_{i=1}^n |w_i| = \norm{1}{\vecw} \eqpun . 
\end{equation}
The formula above can be understood by noting that the worst case for a ReLU unit corresponds to considering an $x$ for which the output is positive, and taking $x'_i = x_i + \epsilon$ if $w_i > 0$, and $x'_i = -\epsilon$ if $w_i < 0$, following essentially the analysis of adversarial examples in \cite{GoodfellowExplainingharnessingadversarial2014}.
Similarly, for a single sigmoid unit with weight vector $\vecw$, as the worst case corresponds to the unit operating in its linear region, we have $s = \frac{1}{4} \norm{1}{\vecw}$, where the factor of $1/4$ corresponds to the maximum derivative of the sigmoid. 
For a \rbfinf\ unit $\neuron(\vecu, \vecw)$, on the other hand, we have:
\begin{equation} \label{eq-sensitivity-rbf}
  s = \frac{2}{e} \cdot \max_{1 \leq i \leq n} u_i^2 = \frac{2}{e} \cdot \norm{\infty}{\vecu}^2 \eqpun . 
\end{equation}
Comparing (\ref{eq-sensitivity-relu}) and (\ref{eq-sensitivity-rbf}), we see that the sensitivity of ReLU and Sigmoid units increases linearly with input size, whereas the sensitivity of \rbfinf\ units is essentially constant with respect to input size. 
This helps understand why attacks that rely on small changes to many inputs work well against networks using ReLU and Sigmoid units, but are not very effective against networks that use \rbfinf\ units, as we will show experimentally in Section~\ref{sec-results-adv}. 

Formulas (\ref{eq-sensitivity-relu}) and (\ref{eq-sensitivity-rbf}) can be extended to bounds for whole networks. 
For a ReLU unit, given upper bounds $\hat{s}_i$ for the sensitivity of input $i$, we can compute an upper bound for the sensitivity of the unit output via
\[
  \hat{s} = \sum_{i=1}^n \hat{s}_i|w_i| \eqpun . 
\]
Of course, this is only an upper bound, as the conditions that maximize the effect on one input will not necessarily maximize the effect on other inputs.
In general, for a ReLU network with $K_0$ inputs and layers of $K_1, K_2, \ldots, K_M$ units, let $W^{(k)} = [w_{ij}]^{(k)}$ be its weight matrices, where $w_{ij}^{(k)}$ is the weight for input $i$ of unit $j$ of layer $k$, for $1 \leq k \leq K_M$.
We can efficiently compute an upper bound $\hat{s}$ for the sensitivity of the network via: 
\begin{equation} \label{eq-sensitivity-network-relu}
  \hat{\vecs}^{(0)} = \vecone \eqpun , \qquad
  \hat{\vecs}^{(k)} = |W^{(k)}| \hat{\vecs}^{(k-1)} \eqpun , \qquad
  \hat{s} = \norm{\infty}{\hat{\vecs}^{(M)}} \eqpun . 
\end{equation}
The formula for Sigmoid networks is identical except for the $1/4$ factors.
Using similar notation, for \rbfinf\ networks we have: 
\begin{equation} \label{eq-sensitivity-network-rbf}
  \hat{\vecs}^{(0)} = \vecone \eqpun , \qquad
  \hat{s}^{(k)}_j = \frac{2}{e} \cdot \max_{1 \leq i \leq K_{k-1}} \hat{s}_i^{(k-1)} 
    \left( u_{ij}^{(k)} \right)^2 \eqpun , \qquad
  \hat{s} = \norm{\infty}{\hat{\vecs}^{(M)}} \eqpun . 
\end{equation}
The interest in (\ref{eq-sensitivity-network-relu}) and (\ref{eq-sensitivity-network-rbf}) is not due to the fact that these formulas provide an accurate characterization of network sensitivity to adversarial attacks. 
They do not, and we will evaluate performance under adversarial attacks experimentally in Section~\ref{sec-results-adv}. 
Rather, by connecting in a simple way the sensitivity to attacks to the network weights, these formulas suggest the possibility of using weight regularization to achieve robustness: by adding $c \hat{s}$ to the loss function for $c > 0$, we might be able to train networks that are both accurate and robust to attacks. 
We will show in Section~\ref{subsec-regularization} that such a regularization helps train more robust \rbfinf\ networks, but it does not help train more robust ReLU networks. 
Incentives do not replace structure, at least in this case.

\section{Training \rbfinf\ Networks}

The non-linearities in (\ref{eq-rbfinf}) make neural networks containing \rbfinf\ units difficult to train using standard gradient descent. 
Indeed, we will show that on permutation-invariant MNIST, that is, on a version of MNIST in which each $28 \times 28$ pixel image is flattened into a $784$-long feature array, training the network using gradient descent yields at most about about 85\% accuracy (see Section~\ref{subsec-pseudo-vs-regular}). 
The problem lies in the shape of RBF functions.
Far from its peak for $\vecx = \vecw$, a function of the form (\ref{eq-rbfinf}) is rather flat, and its derivative may not be large enough to cause the vector of weights $\vecw$ to move towards useful places in the input space during training. 

The solution we have found consists in performing gradient descent using not the true gradient of the function, but rather, using a gradient computed using {\em pseudoderivatives.} 
These {\em pseudoderivatives\/} are functions we use in the chain-rule computation of the loss gradient in lieu of the true derivatives, and are shaped in a way that facilitates learning. 
Employing such pseudoderivatives, our \rbfinf\ networks will be easy to train to high levels of accuracy. 

\subsection{Pseudogradients}

In order to back-propagate the loss gradient through (\ref{eq-rbfinf}), 
indicating $y = \neuron_\infty(\vecu, \vecw)(\vecx)$, we need to compute the partial derivatives $\partial y / \partial x_i$, $\partial y / \partial w_i$, and $\partial y / \partial u_i$, for $1 \leq i \leq n$. 
The true partial derivatives are computed, of course, applying the chain rule of derivation to (\ref{eq-rbfinf}). 
To obtain networks that are easy to train, we replace the derivatives for $\exp$ and $\max$ with alternate functions, which we call {\em pseudoderivatives.}
In the computation of the loss gradient, we will use these pseudoderivatives in place of the true derivatives, yielding a {\em pseudogradient.}

\begin{figure}
\centering
\includegraphics[width=0.5\textwidth]{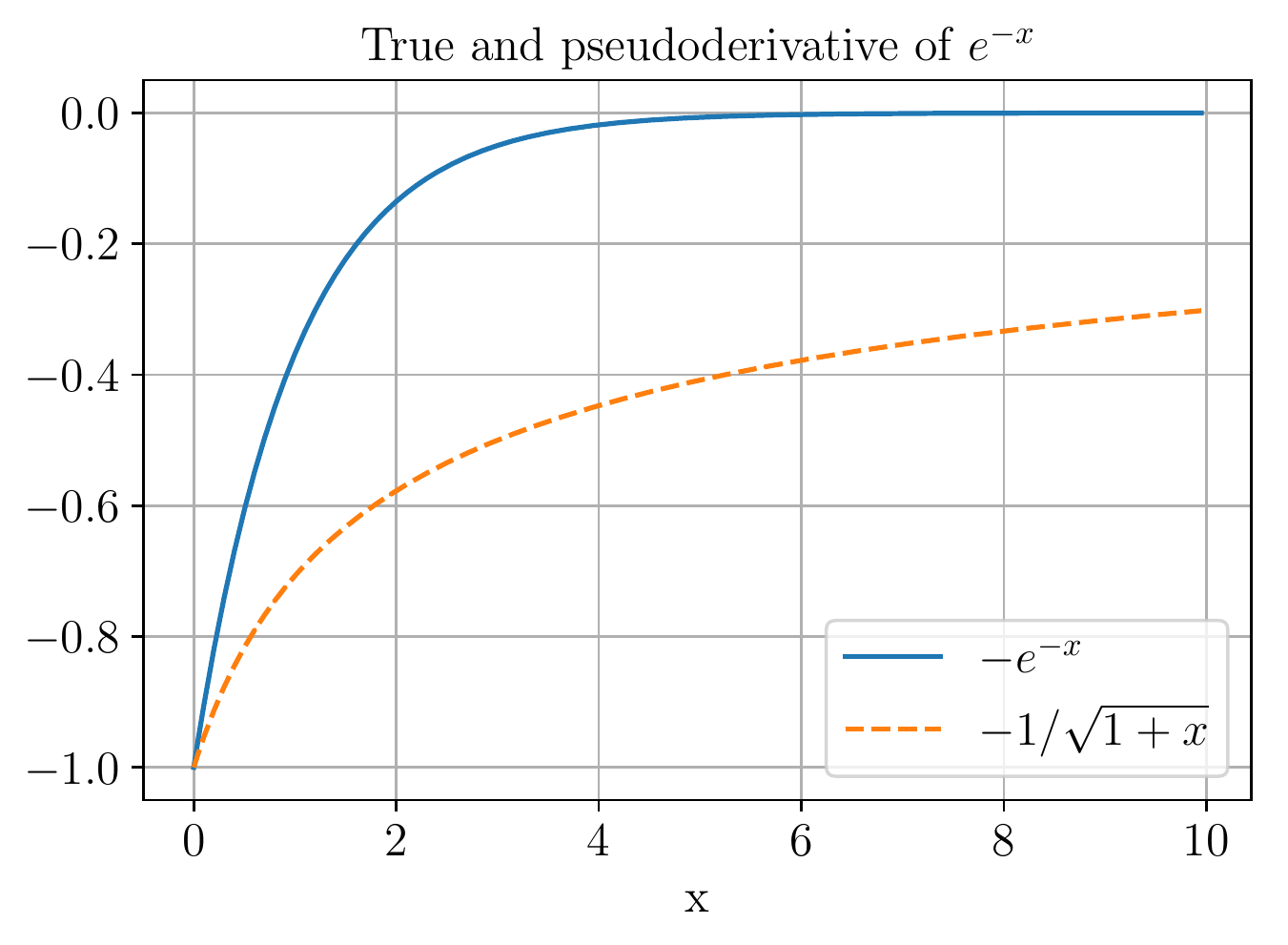}
\caption{True derivative of $e^{-x}$, and pseudoderivative used in training.}
\label{fig-deriv-compare}
\end{figure}

\paragraph{Exponential function.}
In computing the partial derivatives via the chain rule, the first step consists in computing
$
\frac{d}{dz} e^{-z}
$,
which is of course equal to $-e^{-z}$.
The problem is that this derivative is quite small when $z$ is large, and $z$ in (\ref{eq-rbfinf}) is the square of the infinity norm of the scaled distance between $\vecx$ and $\vecw$, which can be large. 
Hence, in the chain-rule computation of the gradient, we replace the true derivative $-e^{-z}$ with the alternate ``pseudoderivative'' below:
\begin{equation} \label{eq-pseudo-exp}
- \frac{1}{\sqrt{1 + z}} \eqpun . 
\end{equation}
The shape of the true and pseudoderivative is compared in Figure~\ref{fig-deriv-compare}. 
As $z$ increases, the pseudoderivative decreases much more slowly than the true derivative.
We experimented with using $- (1 + z)^{-\alpha}$ as pseudoderivative, and we found that it works well for values of $\alpha$ between 0.2 and 0.8; we settled on $\alpha = 1/2$, yielding (\ref{eq-pseudo-exp}).

\paragraph{Max.}
The gradient of $y = \max_{1 \leq i \leq n} z_i$, of course, is given by:
\begin{equation} \label{eq-max-deriv}
\frac{\partial y}{\partial z_i} = 
\begin{cases} 1 & \text{if $z_i = y$} \\ 0 & \text{otherwise.} \end{cases}
\end{equation}
In training, the problem with (\ref{eq-max-deriv}) is that it transmits feedback only to the largest of the inputs to $\max$. 
We found it profitable to transmit feedback not only to the largest input, but also to those that are close in value to it.  
Hence, we obtain the pseudogradient for $\max$ by replacing the right-hand side of (\ref{eq-max-deriv}) with:
\begin{equation} \label{eq-shared-feedback-max}
  e^{z_i - y} \eqpun . 
\end{equation}
In this way, some of the feedback is transmitted to inputs $z_i$ that approach $y$. 
Again, we experimented with functions of the form $e^{\beta(z_i - y)}$, and we found that we could train networks effectively for many values of $\beta$; we settled on $\beta = 1$ for simplicity.

\medskip

One may be concerned that by using the loss pseudogradient as the basis of optimization, rather than the true loss gradient, we may then take optimization steps that cause an increase in true loss. 
This may occur, of course --- but so it may when the amplitude and direction of steps is influenced by algorithms such as AdaDelta \cite{ZeilerADADELTAadaptivelearning2012}, or even gradient descent with momentum. 
Further, the pseudogradient can be zero even when the true gradient is not, and this would cause the training to settle in a position that is not a (local) minimum or saddle point of the cost function. 
However, in practice this is exceedingly rare, and it is a much smaller risk than simply ending up in a suboptimal local minimum.
Ultimately, the efficacy of these training strategies is best evaluated experimentally.

\subsection{Bounding the Weight Range}

In training \rbfinf\ networks, it is useful to bound the range of the $\vecu$ and $\vecw$ weight vectors. 
If no lower bound is used for the individual components $u_i$ of $\vecu$, gradient descent with discrete step sizes can cause the components to become negative, which is useless, as the sign of $u_i$ does not matter, and even counterproductive, as an impulse to lower the value of $u_i$ could result in an overshoot, ending up with an updated value $u'_i < 0$ with $- u'_i > u_i$. 
The upper bound to the components of $\vecu$ performs an important role, as it bounds the sensitivity of the unit outputs with respect to changes in their inputs, as the slope of (\ref{eq-rbfinf}) with respect to the inputs depends on the magnitude of $\vecu$. 

For the first network layer, the weights $w$ should be chosen with a range that matches the input range. 
For subsequent layers, we found that bounding the components of $w$ to the interval $[0, 1]$, which is the output range of a \rbfinf\ unit, works well. 

%% file: adversaries.tex
\section{Generating Adversarial Examples}

We will evaluate the robustness of neural networks with respect to both adversarial attacks, and input noise. 
Consider a network trained with cost function $J(\theta, \vecx, \vecy)$, where $\theta$ is the set of network parameters, $\vecx$ is the input, and $\vecy$ is the output. 
Indicate with $\nabla_\vecx J(\theta, \vecx', \vecy)$ the gradient of $J$ wrt its input $\vecx$ computed at values $\vecx'$ of the inputs, parameters $\theta$, and output $\vecy$.
For each input $\vecx$ belonging to the testing set, given a perturbation amount $\epsilon > 0$, we produce an adversarial example $\advvecx$ with $\norm{\infty}{x - \advvecx} \leq \epsilon$ using the following techniques. 
\paragraph{Fast Gradient Sign Method (FGSM)} \cite{GoodfellowExplainingharnessingadversarial2014}. 
If the cost were linear around $\vecx$, the optimal $\epsilon$-max-norm perturbation of the input would be given by $\epsilon \, \sign(\nabla_\vecx J(\theta, \vecx, \vecy))$. 
This suggests taking as adversarial example: 
\begin{equation} \label{eq-FGSM}
  \advvecx = \clamp{0}{1}{\vecx + \epsilon \, \sign(\nabla_\vecx J(\theta, \vecx, \vecy))} \eqpun , 
\end{equation}
where $\clamp{a}{b}{\vecx}$ is the result of clamping each component of $\vecx$ to the range $[a, b]$; the clamping is necessary to generate a valid MNIST image.

\paragraph{Iterated Fast Gradient Sign Method (I-FGSM)} \cite{KurakinAdversarialexamplesphysical2016}.
Instead of computing a single perturbation of size $\epsilon$ using the sign of the gradient, we apply $M$ perturbations of size $\epsilon/M$, each computed from the endpoint of the previous one. 
Precisely, the attack computes a sequence $\advvecx_0, \advvecx_1, \ldots, \advvecx_M$, where $\advvecx_0 = \vecx$, and where each $\advvecx_{i+1}$ is obtained, for $0 \leq i < M$, by: 
\begin{equation} \label{eq-I-FGSM}
  \advvecx_{i+1} = \clamp{0}{1}{\advvecx_i + \frac{\epsilon}{M} \, \sign(\nabla_\vecx J(\theta, \advvecx_i, \vecy))} \eqpun .
\end{equation}
We then take $\advvecx = \advvecx_M$ as our adversarial example. 
This attack is more powerful than its single-step version, as the direction of the perturbation can better adapt to non-linear cost gradients in the neighborhood of $\vecx$ \cite{KurakinAdversarialexamplesphysical2016}. 

\paragraph{Projected Gradient Descent (PGD)} \cite{MadryDeepLearningModels2017}.
For an input $\vecx \in \reals^n$ and a given maximum perturbation size $\epsilon > 0$, we consider the set $B_\epsilon(\vecx) \inters [0,1]^n$ of valid inputs around $\vecx$, and we perform projected gradient descent (PGD) in $B_\epsilon(\vecx) \inters [0,1]^n$ of the negative loss with which the network has been trained (or, equivalently, projected gradient ascent wrt.\ the loss). 
By following the gradient in the direction of increasing loss, we aim at finding mis-classified inputs in $B_\epsilon(\vecx) \inters [0,1]^n$. 
As the gradient is non-linear, to check for the existence of adversarial attacks we perform the descent multiple times, each time starting from a point of $B_\epsilon(\vecx) \inters [0,1]^n$ chosen uniformly at random. 

\paragraph{Noise.}
In addition to the above adversarial examples, we will study the robustness of our networks by feeding them inputs affected by noise. 
For a testing input $\vecx$ and a noise amount $\epsilon \in [0, 1]$, we produce an $\epsilon$-noisy version $\noisyx$ via 
\[
  \noisyx = (1 - \epsilon) \vecx + \epsilon \noisex \eqpun , 
\]
where $\noisex$ is a random element of the input space. 
For permutation-invariant MNIST, the vector $\noisex$ is chosen uniformly at random in the input space $[0, 1]^n$. 
This noise model is tailored to images: as $\epsilon$ grows, the resulting $\noisyx$ contains less signal, and more white noise. 
Of course, this is one of a multitude of reasonable noise models; nevertheless, it will provide at least a coarse indication of how the networks perform in presence of input perturbations.

\medskip
FGSM and I-FGSM attacks are powerful heuristics for finding adversarial examples, but they are not general search procedures. 
Indeed, \cite{Carlinievaluatingrobustnessneural2017} shows how many networks that resist these attacks still misclassify some adversarial examples. 
Hence, \cite{Carlinievaluatingrobustnessneural2017} argues persuasively that a proper evaluation of susceptibility to adversarial attacks can only be done via general optimization techniques, which search for small input perturbations that cause misclassifications. 
The PGD attacks we have included in our experiments perform a general search and optimization of perturbations with high loss, which can lead to mis-classified examples. 
As in this paper we consider the infinity norm, these attacks are of generality and power comparable with the attacks  described in  \cite{Carlinievaluatingrobustnessneural2017}.

%% file: experiments.tex
\section{Experiments on Permutation-Invariant MNIST}

\subsection{Implementation of \rbfinf\ Networks}

We implemented \rbfinf\ networks in the PyTorch framework \cite{PaszkeAutomaticdifferentiationpytorch2017}; the code is available at \url{\codeurl}.
The resulting implementation is well-suited to running on GPUs, which yield a very large speedup.

PyTorch provides automatic gradient propagation: one needs only define the forward path the feature vectors take in the networks, and the framework takes care of computing and propagating the gradients via a facility called AutoGrad \cite{PaszkeAutomaticdifferentiationpytorch2017}.
Crucially, PyTorch makes it easy to define new functions: for each new function $f$, it is necessary to specify the function behavior $f(\vecx)$, and the function gradient $\grad_\vecx f$.
This makes it very easy to implement \rbfinf\ networks.
All that is required consists in implementing two special functions: a {\em LargeAttractorExp\/} function, which has forward behavior $e^{-x}$ and backward gradient propagation according to $-1/\sqrt{1 + x}$, and {\em SharedFeedbackMax,} which behaves forward like $\max$, and backwards as given by (\ref{eq-shared-feedback-max}). 
These two functions are then used in the definition of \rbfinf\ units, as per (\ref{eq-rbfinf}) and (\ref{eq-rbfinf-or}), with the AutoGrad facility of PyTorch providing then backward gradient propagation for the complete networks.

We also introduced to PyTorch {\em bounded parameters,} which have a prescribed range that is enforced during network training; these bounded parameters were used to implement our $\vecu$ and $\vecw$ weight vectors. 

\subsection{Experimental Setup}

\paragraph{Dataset.}
We use the MNIST dataset \cite{LeCunGradientbasedlearningapplied1998} for our experiments, following the standard setup of 60,000 training examples and 10,000 testing examples. 
Each digit image was flattened to a one-dimensional feature vector of length $28 \times 28 = 784$, and fed to a fully-connected neural network; this is the so-called {\em permutation-invariant\/} MNIST. 

\paragraph{Neural networks.}
We compared the accuracy of the following network structures. 
\begin{itemize}
\item {\bf Sigmoid.}
We consider fully-connected networks consisting of sigmoid units. 
The last layer consists of 10 units, corresponding to the 10 digits. 
We train sigmoid networks using square-error as loss function; we experimented with cross-entropy loss and other losses, and square-error loss performed as well as any other loss in our experiments.

\item {\bf ReLU.}
We consider fully-connected ReLU networks \cite{NairRectifiedlinearunits2010,KrizhevskyImagenetclassificationdeep2012a}, again whose last layer consists of 10 units.
The output of ReLU networks is fed into a softmax, and the network is trained via cross-entropy loss. 

\item {\bf \rbfinf.}
We consider fully-connected networks consisting of \rbfinf\ units. 
As for networks of sigmoid units, the last layer has 10 units, and the network is trained using square-error loss. 
For a \rbfinf\ network with $m$ layers, we denote its type as $\rbfnet(K_1, \ldots, K_m \mid t_1, \ldots, t_m)$, where $K_1, \ldots, K_m$ are the numbers of units in each layer, and where the units in layer $i$ are And units if $t_i = \und$, Or units if $t_i = \oder$, and are a random mix of And and Or units if $t_m = \ast$. 

\end{itemize}
Obviously, a network could mix layers consisting of sigmoid, ReLU, and \rbfinf\ units, or indeed the units could be mixed in each layer, but we have not experimented with such hybrid architectures.
Unless otherwise noted, we use bounds of $[0.01, 3]$ for the components of the $u$-vectors, and $[0, 1]$ for the $w$-vectors, the latter corresponding to the value range of MNIST pixels. 

\paragraph{Training and testing.}
We trained all networks with the AdaDelta optimizer \cite{ZeilerADADELTAadaptivelearning2012}, which yielded good results for all networks considered.  
Unless otherwise noted, we performed 10 runs of each experiment with different seeds for the random generator used for weight initialization, in order to measure the mean and standard deviation of each result. 
When reporting an accuracy as $X\% \pm Y\%$, $X$ is the mean value of the percent accuracy, and $Y$ the standard deviation of the individual run outcomes (rather than the standard deviation of the mean). 
Error bars in the plots correspond to one standard deviation of individual run results.

\paragraph{Attacks.}
We performed the adversarial attacks as follows. 
For FGSM, and for noise, the attack is completely determined by its amplitude $\epsilon$; the attacks were applied to all examples in the test set. 
In I-FGSM attacks, we performed 10 iterations of (\ref{eq-I-FGSM}).  
The attack was applied to all data in the test set. 

As PGD attacks are considerably more computationally intensive than the other attacks considered, for each experiment we select the neural network model trained by the first of our set of runs, and we compute the performance under PGD attacks over the first 5,000 examples in the test set, or half of it. 
To generate adversarial examples for input $\vecx$, we start from a random point in $\ball_\epsilon(\vecx)$ and we perform 100 steps of projected gradient descent using the AdaDelta algorithm to tune step size; if at any step a misclassified example is generated, the attack is considered successful. 
For each input $\vecx$, we repeat this procedure 20 times, that is, our search is performed with 20 restarts. 
We experimented with using more than 100 steps of descent, but we observed only minimal changes in attack success rates. 
We also found that when the network accuracy in presence of PGD attacks was at least 80\%, additional restarts contributed very little to the attack efficacy. 
When the attack succeeded over 20\% of the time, or for networks whose accuracy under attack was below 80\%, additional restarts did increase somewhat the overall attack success, so that our results should be considered upper bounds for classification accuracy.
In any case, as we kept the attack configuration identical for all our experiments, our results will enable us to compare the performance of the different neural network architectures in presence of PGD attacks. 

\subsection{Performance of \rbfinf, ReLU, and Sigmoid Networks}
\label{sec-results-adv}

\begin{figure}[t]
\centering
\includegraphics[width=0.49\textwidth]{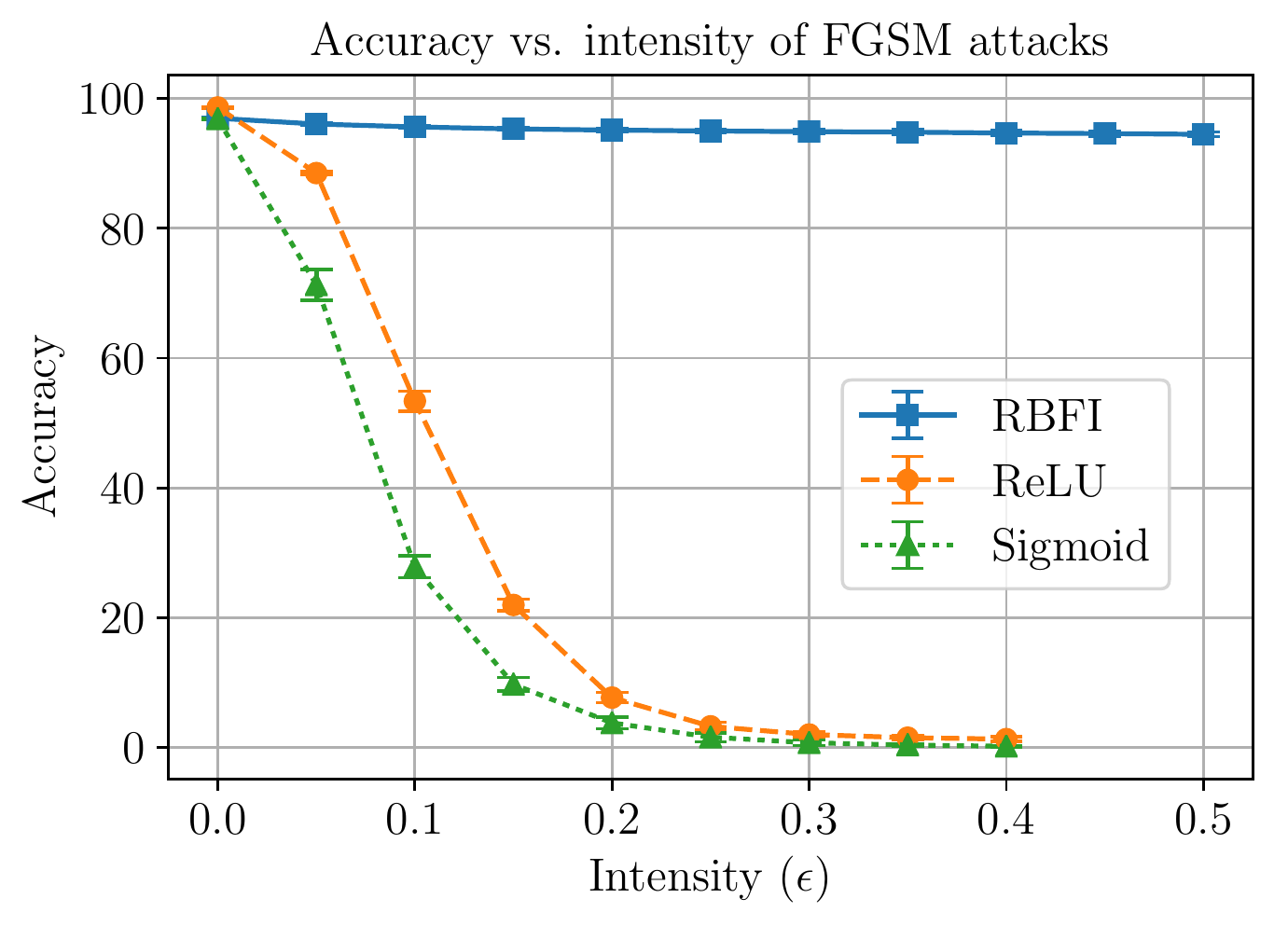} 
\includegraphics[width=0.49\textwidth]{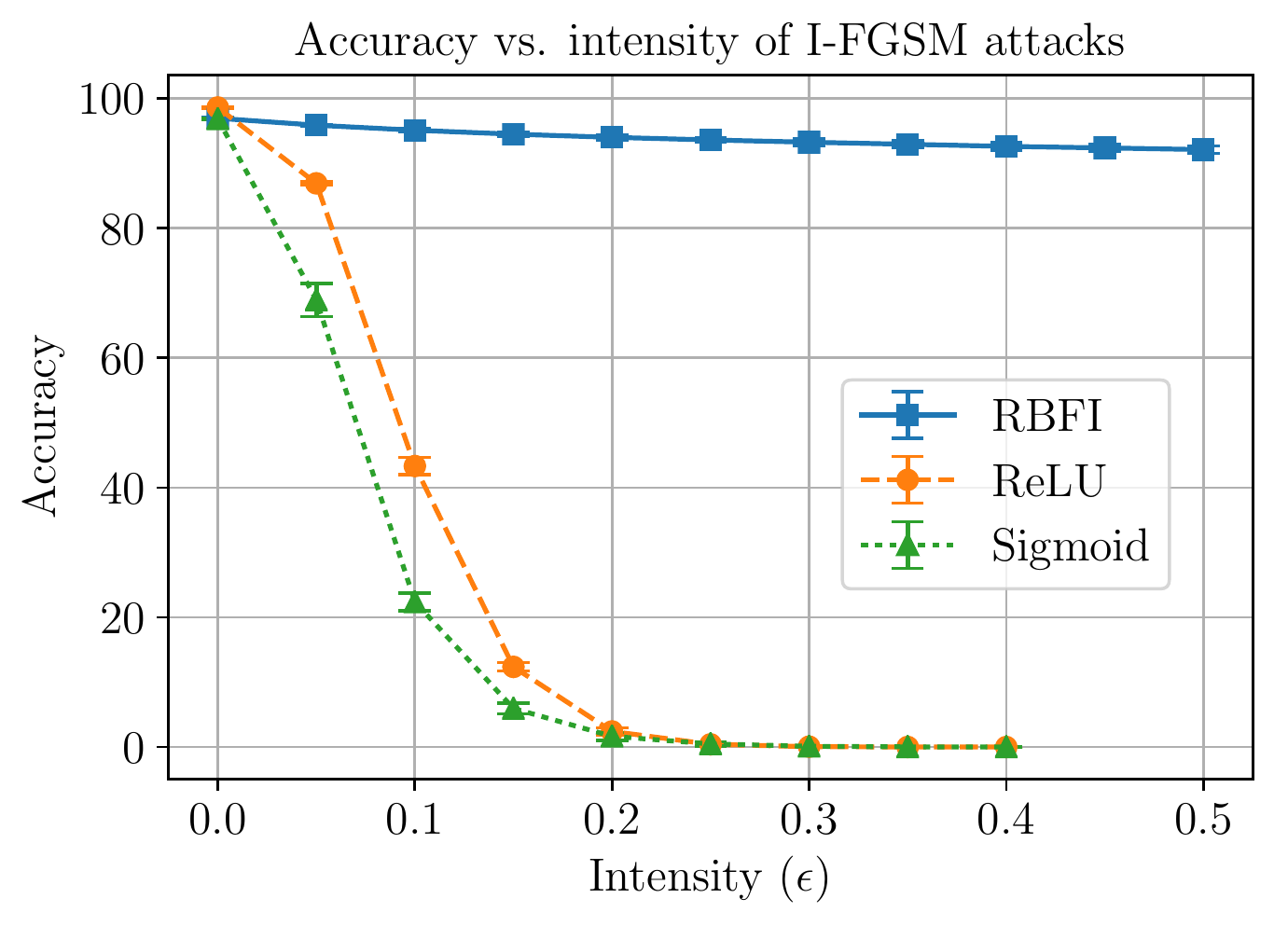} \\
\includegraphics[width=0.49\textwidth]{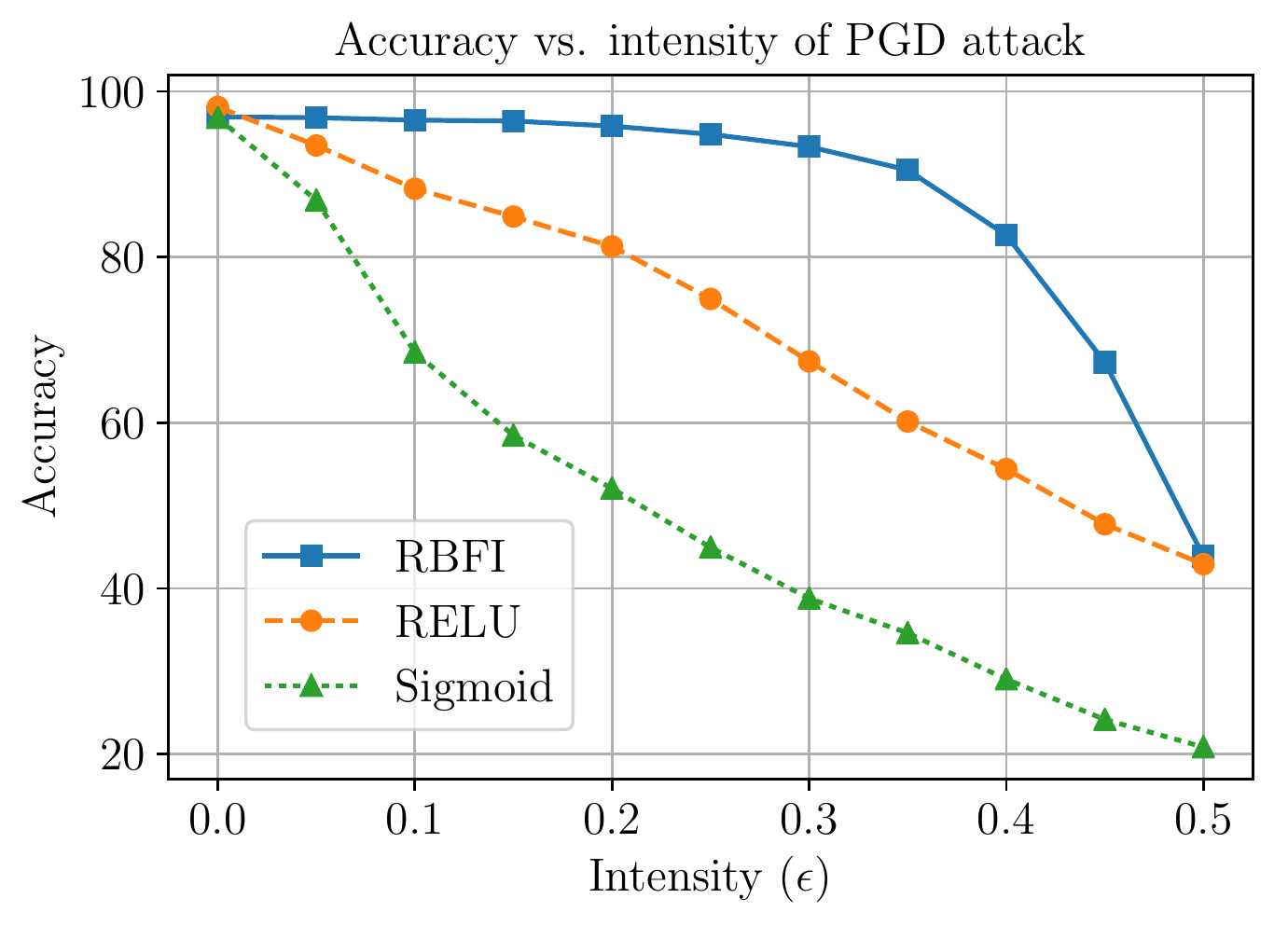}
\includegraphics[width=0.49\textwidth]{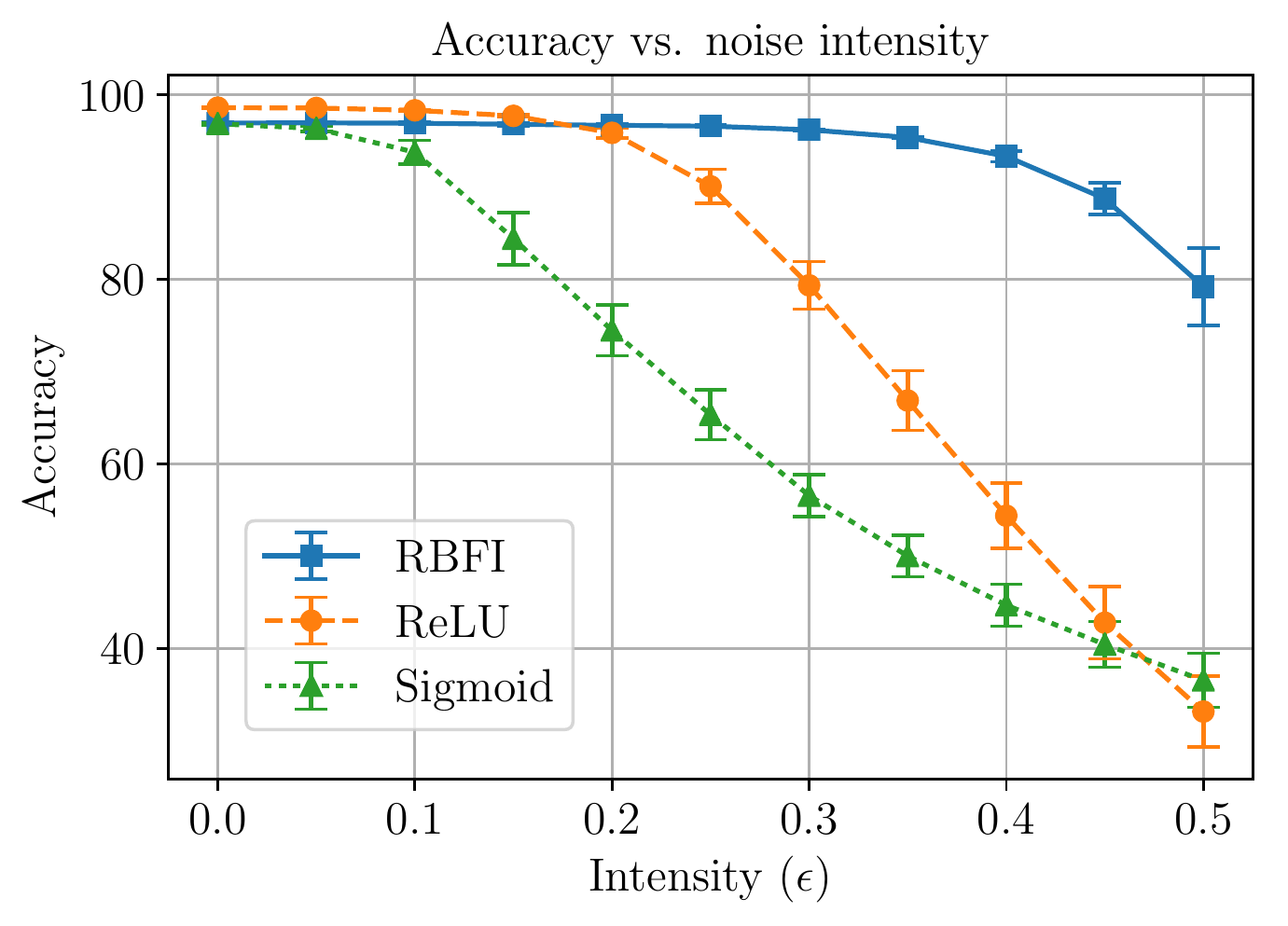}
\caption{Performance of 512-512-512-10 networks in presence of adversarial examples and noise.
The \rbfinf\ network has layers $\rbfnet(512, 512, 512, 10 \mid \und, \oder, \und, \oder)$.}
\label{fig-standard-training}
\end{figure}

\begin{table}[h]
\centering
\begin{tabular}{r||r|r|r|r|r}
Network
 & Accuracy ($\epsilon=0$) 
 & FGSM, $\epsilon = 0.3$ 
 & I-FGSM, $\epsilon = 0.3$
 & PGD, $\epsilon = 0.3$
 & Noise, $\epsilon = 0.3$ \\ \hline
ReLU 
 & $\mathbf{98.62 \pm 0.08}$
 & $1.98 \pm 0.42$
 & $0.06 \pm 0.06$
 & $67.40$
 & $79.36 \pm 2.60$ \\
Sigmoid
 & $96.88 \pm 0.15$
 & $0.71 \pm 0.43$
 & $0.11 \pm 0.11$
 & $38.78$
 & $56.57 \pm 2.28$ \\
\rbfinf
 & $96.96 \pm 0.14$ 
 & $\mathbf{94.90 \pm 0.35}$ 
 & $\mathbf{93.27 \pm 0.48}$ 
 & $\mathbf{93.32}$ 
 & $\mathbf{96.23 \pm 0.08}$ \\
 \end{tabular}
 \caption{Performance of 512-512-512-10 for MNIST testing input, and in presence of adversarial examples and noise computed with perturbation size $\epsilon=0.3$. 
 The \rbfinf\ network has layers $\rbfnet(512, 512, 512, 10 \mid \und, \oder, \und, \oder)$.}
 \label{table-standard-training} 
\end{table}

We first give the results on the accuracy and resistance to adversarial examples for networks trained on the standard MNIST training set, without the benefit of adversarial examples. 
We conducted 10 measurement runs for ReLU and Sigmoid networks, and 5 for \rbfinf\ networks, which suffices to determine result variance.
In each run, we trained the networks for 30 epochs on the MNIST training set, and we evaluated performance both on the original MNIST training set, and on the training set perturbed either by noise, or via adversarial examples computed via FGSM, I-FGSM, or PGD. 
We chose 30 epochs as all networks reached their peak performance by then. 
We give results for 4-layer networks, with 512, 512, 512, and 10 units.
This is a size at which all the network types have essentially reached their peak performance. 
For the \rbfinf\ network we chose geometry $\rbfnet(512, 512, 512, 10 \mid \und, \oder, \und, \oder)$; as we will detail in Section~\ref{subsec-geometry}, using networks with alternating unit types gives marginally better results. 

We report the results in Figure~\ref{fig-standard-training} and in Table~\ref{table-standard-training}. 
In absence of perturbations, \rbfinf\ networks lose $(1.66 \pm 0.21)\%$ performance compared to ReLU networks (from $(98.62\pm0.07)\%$ to $(96.96\pm0.14)\%$), and perform comparably to sigmoid networks (the difference is below the standard deviation of the results).
When perturbations are present, in the form of adversarial attacks or noise, the performance of \rbfinf\ networks is markedly superior.

We note that the FGSM and I-FGSM attacks are not effective against \rbfinf\ networks.  
This phenomenon, observed in \cite{Carlinievaluatingrobustnessneural2017} also for networks trained via defensive distillation \cite{PapernotDistillationdefenseadversarial2016,CarliniDefensivedistillationnot2016}, is likely due to the fact that for \rbfinf\ networks the gradient in proximity of valid inputs offers only limited information about the possible location of adversarial examples. 
Indeed, even noise provides a better adversary than FGSM and I-FGSM. 
The PGD attacks are more powerful, as they explore $B_\epsilon(\vecx) \inters [0,1]^n$ more thoroughly, sampling points in it at random before following the gradient. 

For systems that exhibit marked linearity, such as ReLU and sigmoid networks, FGSM and I-FGSM attacks are instead more powerful than PGD attacks, at least if the latter are carried out with a moderate number of restarts, as in our experiments. 
In FGSM and I-FGSM, we perturb each input coordinate by $\epsilon$, according to the gradient sign; this may lead to a more effective exploration of $B_\epsilon(\vecx) \inters [0,1]^n$ than by following the gradient, which tends to explore more the input coordinates for which the gradient component is larger in absolute value.
Perhaps by using thousands or millions of restarts, PGD attacks could become as effective as FGSM and I-FGSM attacks for ReLU networks. 
In our experiments, going from 20 to 100 restarts did not appreciably increase the effectiveness of PGD attacks for $\epsilon \leq 0.3$.

\subsection{Performance of Networks Trained With Adversarial Examples}
\label{subsec-learning-adversarial}

\begin{figure}
\centering
\includegraphics[width=0.49\textwidth]{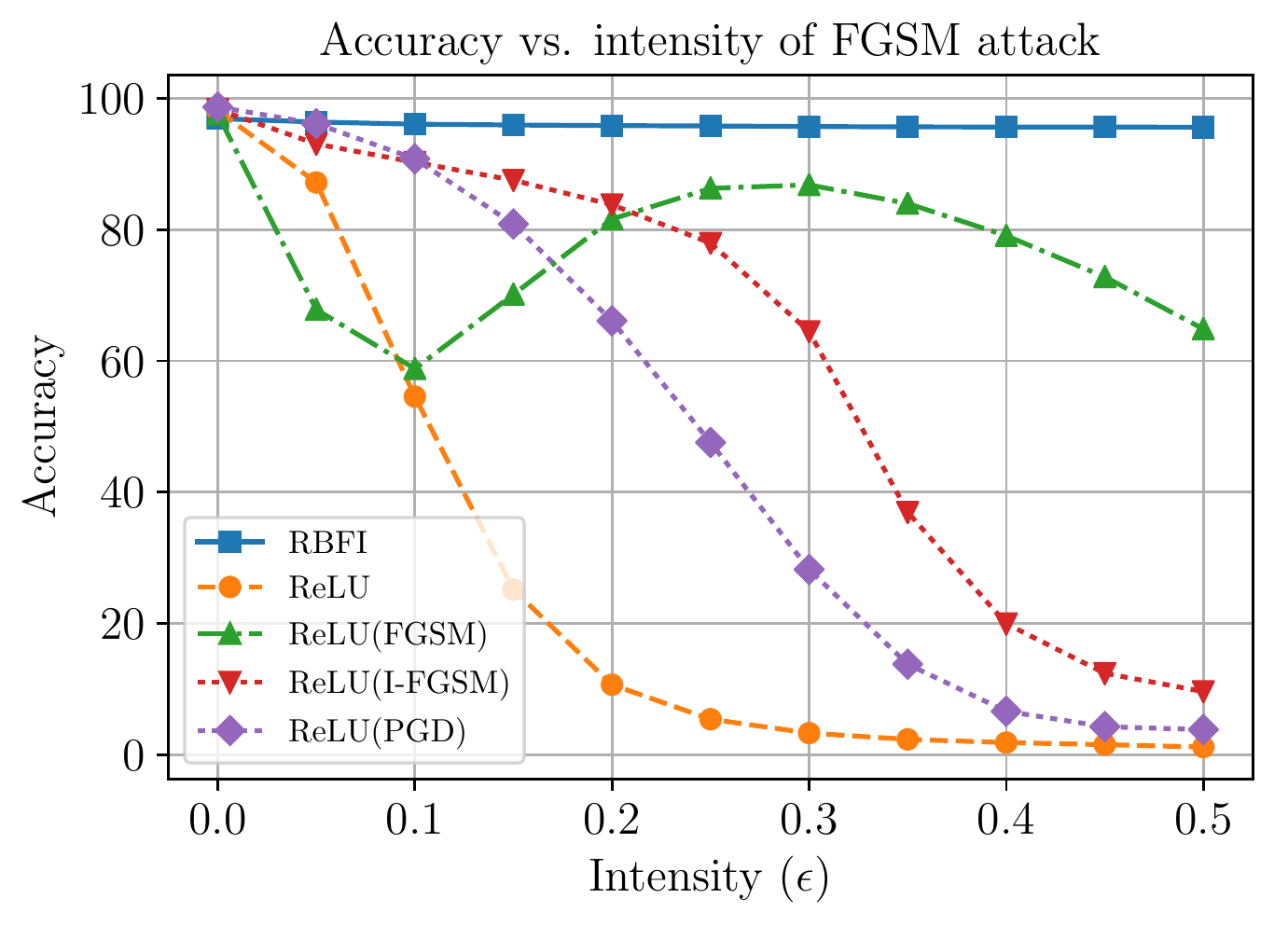}
\includegraphics[width=0.49\textwidth]{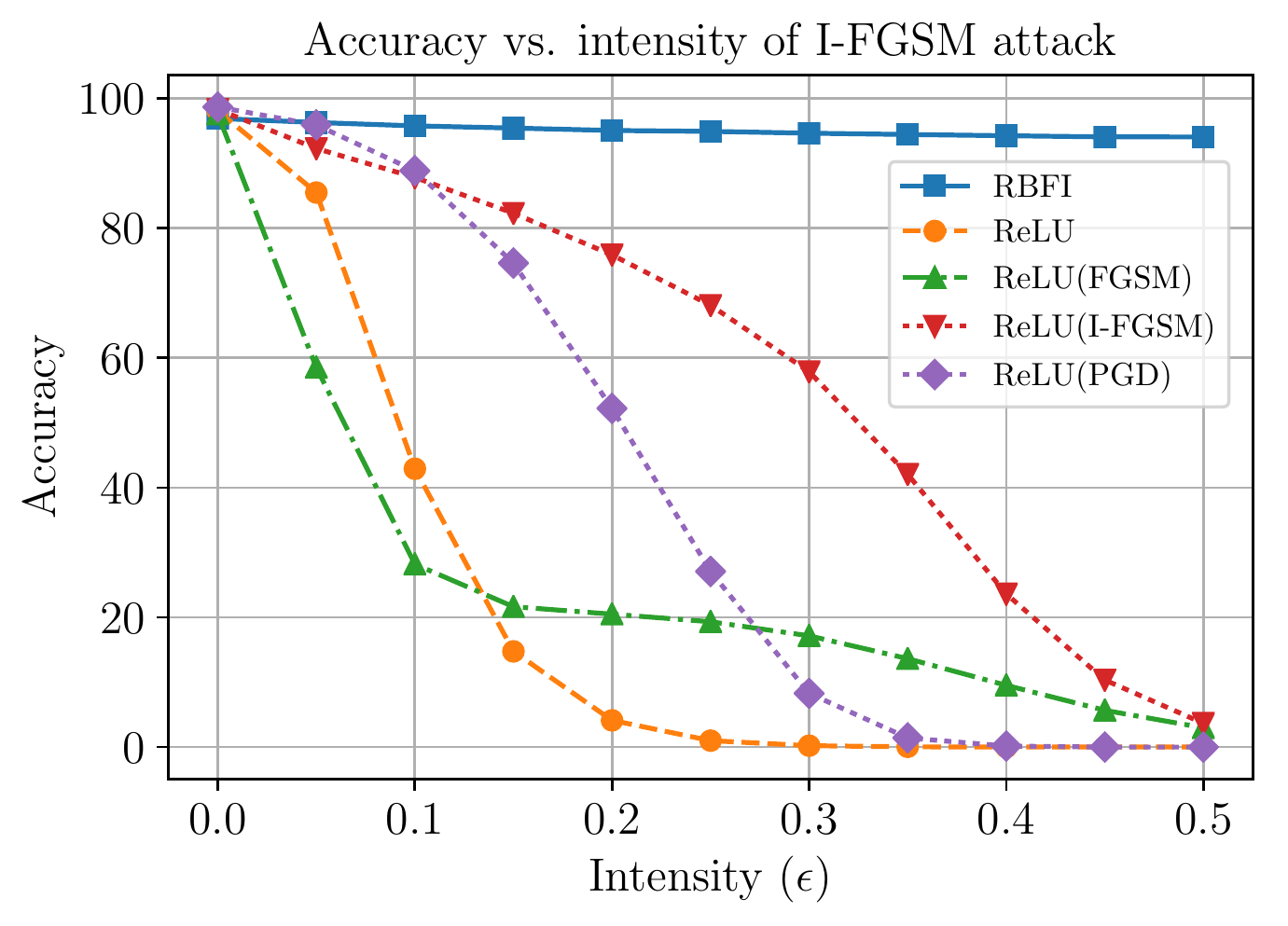}\\
\includegraphics[width=0.49\textwidth]{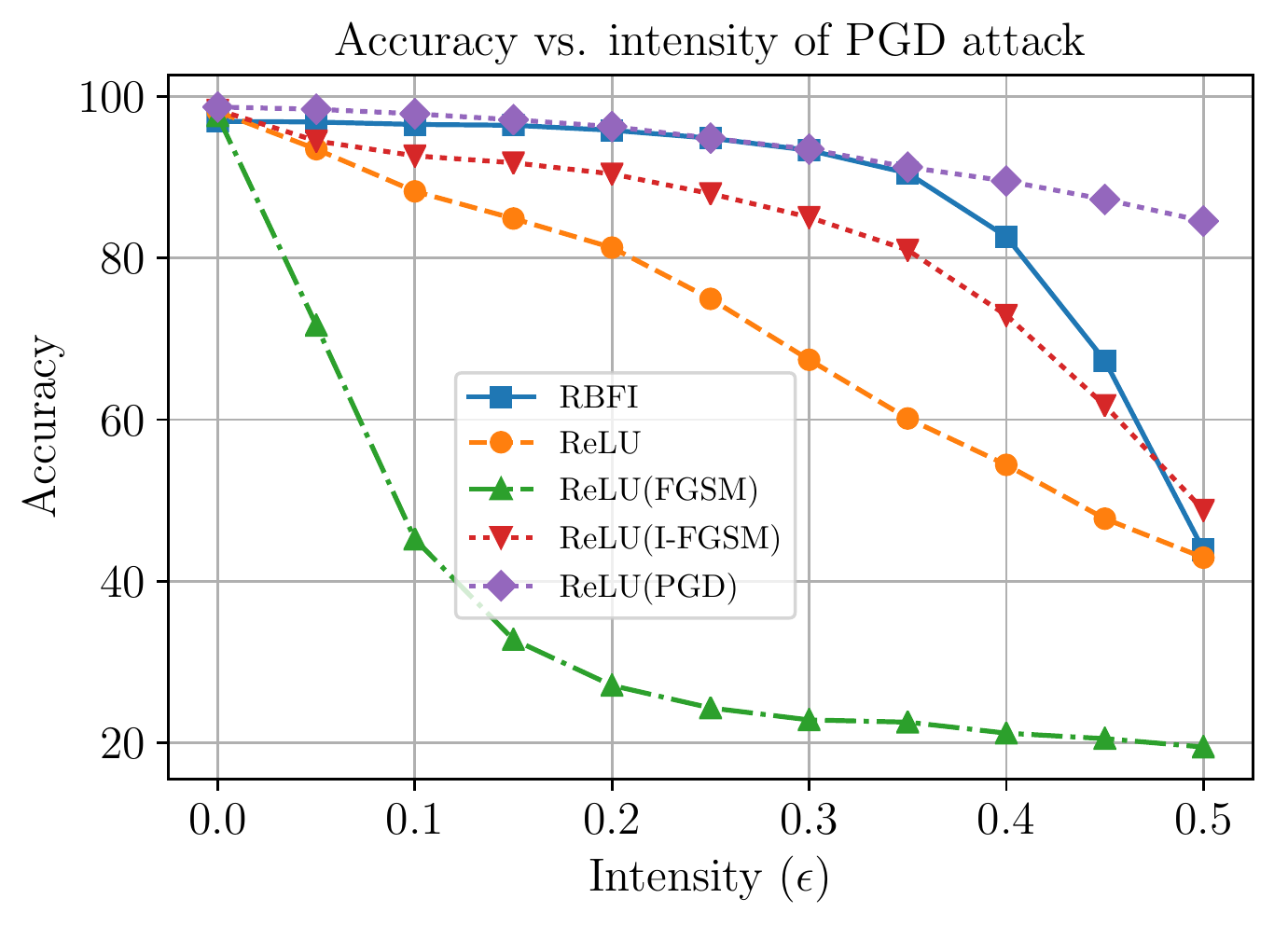}
\includegraphics[width=0.49\textwidth]{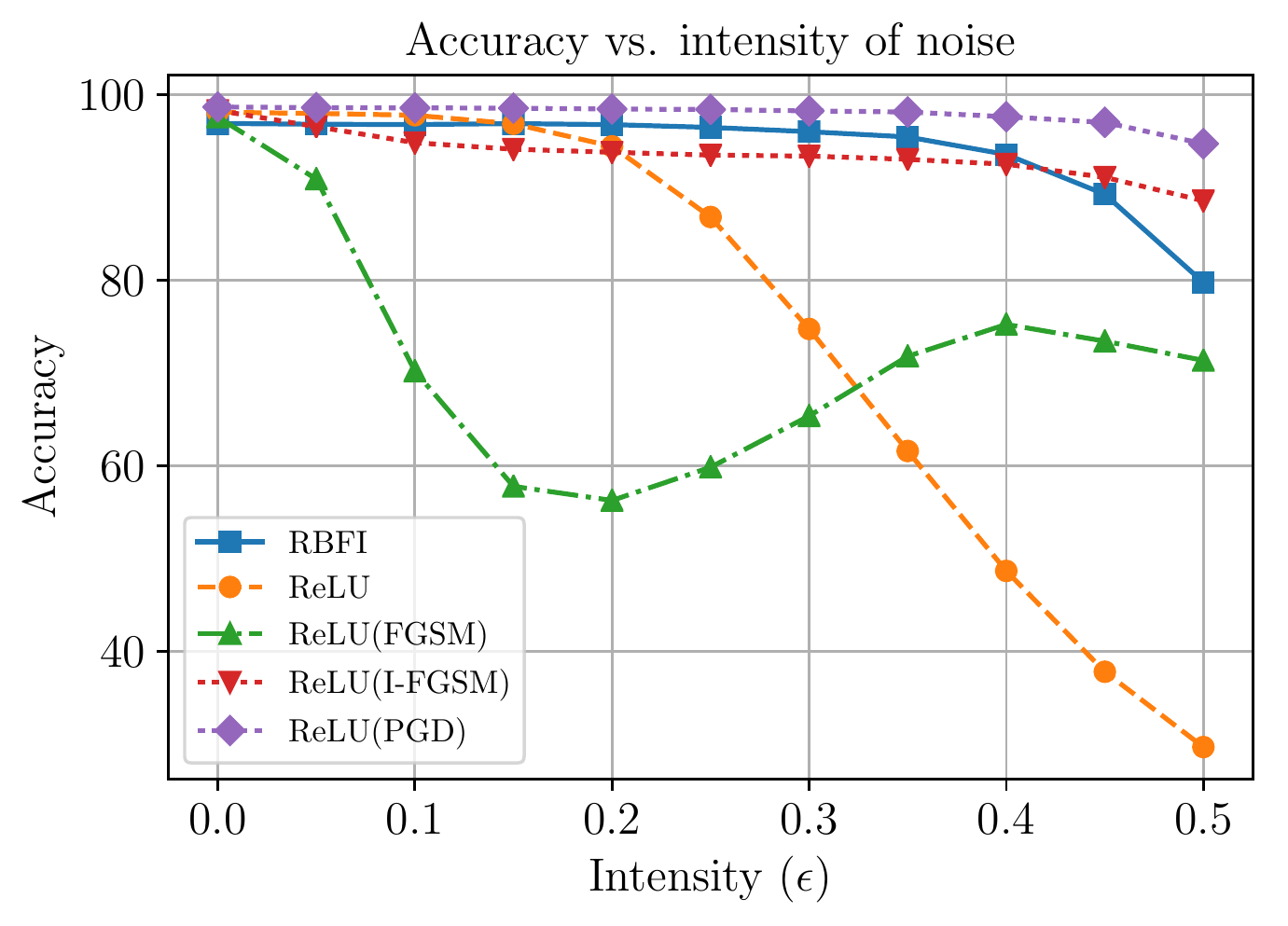}
\caption{Performance of ReLU networks trained with adversarial examples, vs.\ performance of \rbfinf\ network trained normally, with respect to adversarial input and noise.}
\label{fig-adversarial-training}
\end{figure}

Including adversarial examples in the training set is the most common method used to make neural networks more resistant to adversarial attacks \cite{GoodfellowExplainingharnessingadversarial2014,MadryDeepLearningModels2017}.
Here, we explore whether ReLU and Sigmoid networks trained via a mix of normal and adversarial examples can offer a resistance to adversarial attacks compared to that offered by \rbfinf\ networks trained on standard examples only. 
Specifically, we compared the performance of a \rbfinf\ network with that of ReLU network trained for 10 epochs as follows: 
\begin{itemize}
\item {\bf ReLU:} trained normally, on the pairs $(\vecx, t)$ of input $\vecx$ and class $t$ in the standard training set. 

\item {\bf ReLU(FGSM):}
for each $(\vecx, t)$ in the training set, we construct an adversarial example $\advvecx$ via (\ref{eq-FGSM}), and we feed both $(\vecx, t)$ and $(\advvecx, t)$ to the network for training. 

\item {\bf ReLU(I-FGSM):}
for each $(\vecx, t)$ in the training set, we construct an adversarial example $\advvecx$ via (\ref{eq-I-FGSM}), and we feed both $(\vecx, t)$ and $(\advvecx, t)$ to the network for training. 

\item {\bf ReLU(PGD):}
for each $(\vecx, t)$ in the training set, we perform 100 steps of projected gradient descent from a point chosen at random in $B_\epsilon(\vecx) \inters [0,1]^n$; denoting by $\vecx'$ the ending point of the projected gradient descent, we feed both $(\vecx, t)$ and $(\vecx', t)$ to the network for training. 

\end{itemize}
We did not include sigmoid networks in the comparison, as they could be trained only with respect to FGSM-generated adversarial examples, for which they gave inferior results.
We generated adversarial examples for $\epsilon=0.3$, which is consistent with \cite{MadryDeepLearningModels2017}; $\epsilon=0.3$ also and that, with respect to the results reported in Figure~\ref{fig-standard-training} offers a middle value in the range of perturbations we use for testing. 

Obviously, many variations are possible to the above adversarial training regime. 
It is possible to vary the proportion of normal and adversarial inputs. 
It is also possible to run PGD multiple times for every input, choosing the end-point with the highest loss, at a price of increased computational cost (the above PGD training, by requiring 100 steps of gradient descent for each input, already makes training 100 times more expensive).
We could also generate adversarial examples for many values of $\epsilon$, rather than $\epsilon=0.3$ only. 
While the variations are many, we believe the above four attacks enable us to gain at least some understanding of how networks trained in adversarial fashion compare with \rbfinf-based networks. 

As in the previous section, we use networks with 512-512-512-10 units, and we choose geometry $\rbfnet(512, 512, 512, 10 \mid \und, \oder, \und, \oder)$ for the \rbfinf\ network. 
Training was performed for 10 epochs for Sigmoid and ReLU networks, which seemed sufficient. 
The results are given in Figure~\ref{fig-adversarial-training}.
Overall, the best networks may be the simple \rbfinf\ networks, trained without the use of adversarial examples: for each class of attack, they exhibit either the best performance, or they are very close in performance to the best performer; this is true for no other network type. 
For PGD attacks, the best performance is obtained via ReLU(PGD) networks, trained on PGD attacks; however, ReLU(PGD) networks do not fare well in presence of FGSM or I-FGSM attacks.  

We note that ReLU(FGSM) networks seem to learn that $\epsilon=0.3$ FGSM attacks are likely, but they have not usefully generalized the lesson, for instance, to attacks of size $0.1$. 
The S-shaped performance curve of ReLU(FGSM) with respect to FGSM or noise is known as {\em label leaking:\/} the network learns to recognize the original input given its perturbed version \cite{KurakinAdversarialmachinelearning2016}.

\subsection{\rbfinf\ Networks with And, Or, and Mixed Layers}
\label{subsec-geometry}

\begin{table}
\centering
\begin{tabular}{r|r||r|r||r|r}
    \multicolumn{2}{c||}{$\rbfnet(128,128,10)$} & 
    \multicolumn{2}{c||}{$\rbfnet(64,64,64,10)$} &
    \multicolumn{2}{c}{$\rbfnet(512,512,512,10)$} \\
    Layers & Accuracy & Layers & Accuracy \\ \hline
$\ast,\ast,\oder$  & $95.00 \pm 0.29$ & $\ast,\ast,\ast,\oder$ & $93.64 \pm 0.30$ & $\ast,\ast,\ast,\oder$ & $96.79 \pm 0.17$ \\ 
$\ast,\ast,\und $  & $94.22 \pm 0.30$ & $\ast,\ast,\ast,\und$ & $93.66 \pm 0.34$ & $\ast,\ast,\ast,\und$ & $96.87 \pm 0.22$ \\
$\oder,\und,\oder$ & $94.94 \pm 0.21$ & $\und,\oder,\und,\oder$ & $93.46 \pm 0.45$ & $\und,\oder,\und,\oder$ & $96.96 \pm 0.14$ \\
$\und,\oder,\und$  & $94.26 \pm 0.32$ & $\oder,\und,\oder,\und$ & $93.64 \pm 0.23$ \\
$\und,\und,\und$   & $94.25 \pm 0.21$ & $\und,\und,\und,\und$ & $93.30 \pm 0.49$ \\
& & $\oder,\oder,\oder,\oder$ & $93.69 \pm 0.24$ \\
\end{tabular}
\caption{Accuracy of networks trained with different layer kinds.
The $128,128,10$ and $512,512,512,10$ networks were trained for 30 epochs; the $64,64,64,10$ networks were trained for 10 epochs.}
\label{table-geometry}
\end{table}

Logic circuits are easy to design in alternating And-Or layers (with the possibility of complementing gate inputs, which \rbfinf\ units allow). 
Hence, we expected networks with alternating unit types to offer superior performance. 
The results of our experiments with layer types are presented in Table~\ref{table-geometry}.
We see that for small network sizes, there are some differences, but the differences generally become statistically insignificant (smaller than the standard deviation in the run results) as the networks become larger.
We do not yet have a good insight in why accuracy differences are not larger.

\subsection{Pseudogradients vs.\ Standard Gradients}
\label{subsec-pseudo-vs-regular}

\begin{table}
\centering
\begin{tabular}{l||r|r}
 & \multicolumn{2}{c}{Accuracy} \\
 Network & Regular Gradient & Pseudogradient \\ \hline
 $\rbfnet(512,512,512,10 \mid \ast,\ast,\ast,\vee)$ & $86.35 \pm 0.75$ & $96.79 \pm 0.17$ \\
 $\rbfnet(256,256,256,10 \mid \ast,\ast,\ast,\vee)$ & $85.12 \pm 1.07$ & $96.63 \pm 0.16$ \\
 $\rbfnet(128,128,128,10 \mid \ast,\ast,\ast,\vee)$ & $83.25 \pm 1.34$ & $95.80 \pm 0.20$ \\
 $\rbfnet(256,256,10 \mid \ast,\ast,\vee)$ & $82.94 \pm 1.64$ & $95.60 \pm 0.25$ \\
 $\rbfnet(128,128,10 \mid \ast,\ast,\vee)$ & $82.40 \pm 3.72$ & $95.00 \pm 0.29$
 \end{tabular}
\caption{Accuracy achieved training \rbfinf\ networks over 30 epochs using regular gradients and pseudogradients.}
\label{table-regular-derivatives}
\end{table}

In order to justify the somewhat un-orthodox use of pseudogradients, we compare the performance achieved by training \rbfinf\ networks with standard gradients, and with pseudogradients. 
Table~\ref{table-regular-derivatives} compares the accuracy attained training \rbfinf\ networks using regular gradients, and pseudogradients. 
We see that training with the true gradient yielded markedly inferior results. 
The training with regular gradients is also slower to converge; the accuracy in Table~\ref{table-regular-derivatives} was measured after 30 training epochs, when the accuracy of training with both regular, and pseudogradients, had plateaued.

\subsection{Learning with Regularization}
\label{subsec-regularization}

\begin{figure}
\centering
\includegraphics[width=0.49\textwidth]{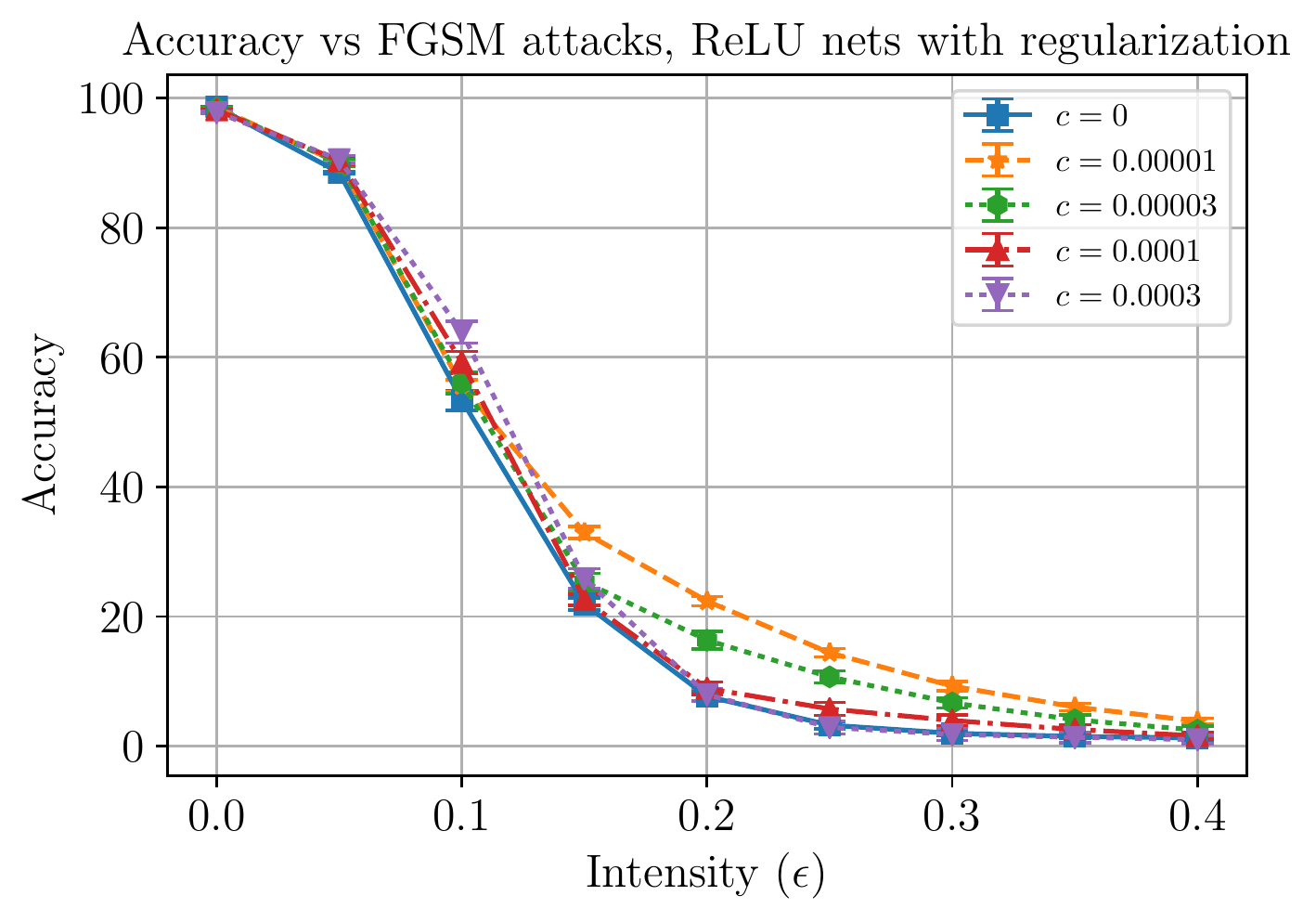}
\includegraphics[width=0.49\textwidth]{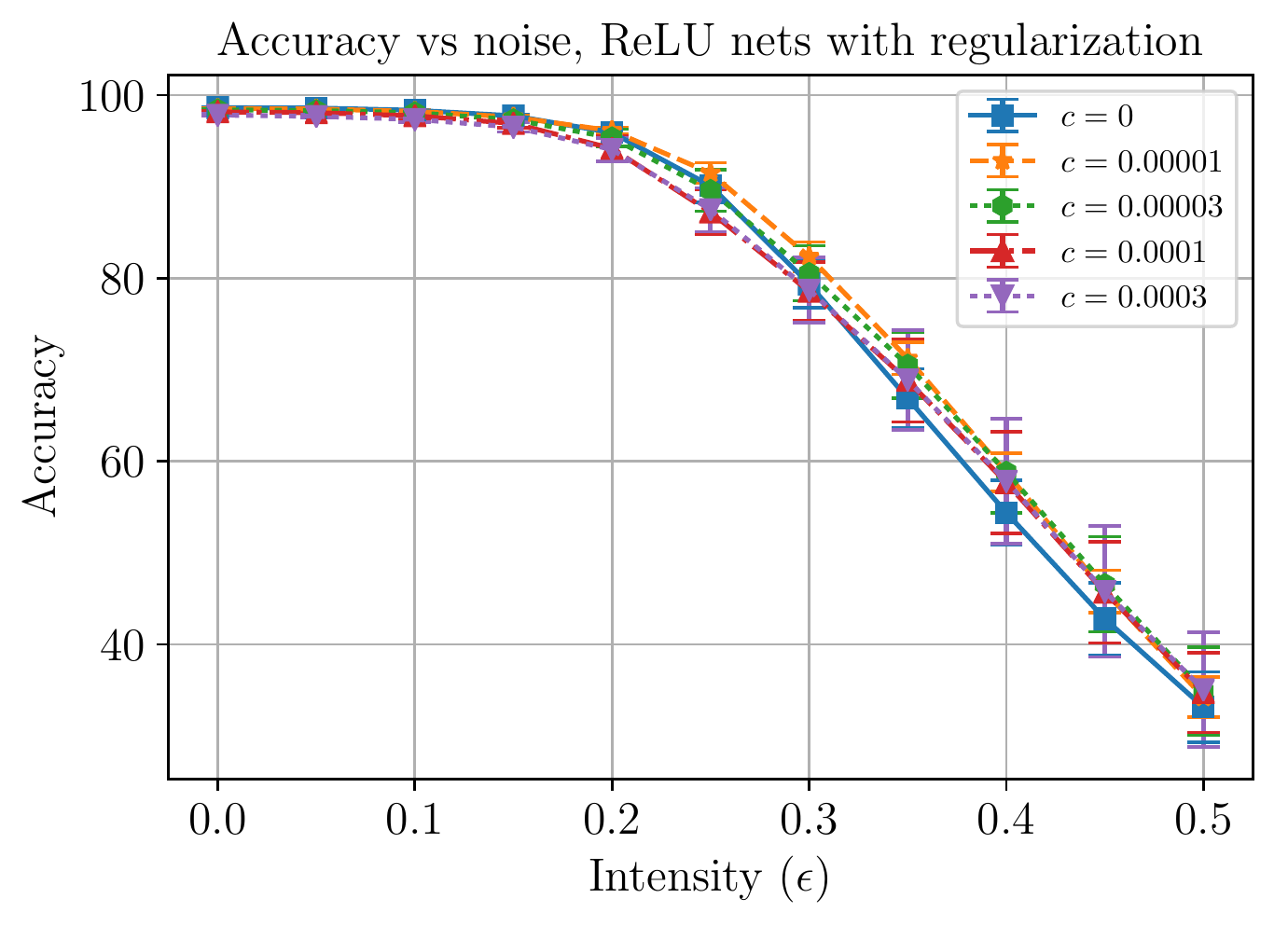}
\caption{Performance of 512-512-512-10 ReLU networks trained for 30 epochs with an amount $c$ of regularization.  
For $c \geq 0.001$, the ReLU network did not learn (performed no better than a random guess).}
\label{fig-regularization}
\end{figure}

\paragraph{ReLU networks.}
In this paper we suggest that neural network may be made more resistant to adversarial attacks via an appropriate choice of structure, namely, by using \rbfinf\ units. 
One may wonder whether the same effect could be achieved more simply by employing weight regularization and standard ReLU units.
In Section~\ref{subsec-sensitivity} we described how to obtain bounds $\hat{s}$ for attack sensitivity in terms of network weights, via (\ref{eq-sensitivity-network-relu}) or (\ref{eq-sensitivity-network-relu}). 
It is thus natural to consider using $\hat{s}$ as regularization, adding to the loss used to train the networks $c \hat{s}$, for $c \geq 0$. 
The results of doing so for ReLU networks is illustrated in Figure~\ref{fig-regularization}.
Moderate values of $c$ yield small improvements, and larger values of $c$ cause the network to cease learning, as keeping the weights low to minimize $c\hat{s}$ takes the precedence over reducing the accuracy-related component of the loss. 
In summary, at least in our attempts, regularization could not replace structure. 

\paragraph{\rbfinf\ networks.}
The choice of upper bound for the components of the $\vecu$-vector influences the resistance of the trained networks to adversarial examples, as indicated by (\ref{eq-sensitivity-rbf}) and from the bound (\ref{eq-sensitivity-network-rbf}). 
MNIST is a digit recognition dataset: pixels that do not contain digit portions are close to 0 in value, while pixels that contain digit portions tend to be close to 1. 
This provides some heuristic for choosing the bound for the components of $\vecu$: our chosen bound of 3 enables to differentiate between pixel value $0$ and $0.3$ by an amount $e^0 - e^{-(3 \cdot 0.3)^2} \approx 0.56$, and this seems sufficient sensitivity; indeed, the results of Section~\ref{sec-results-adv} confirm this.
We may ask: would \rbfinf\ networks perform as well in settings where we do not have a good intuition for an upper bound for $\vecu$?
The answer is yes, provided they are trained with a regularization that provides an incentive towards low $\vecu$ values.

\begin{figure}
\centering
\includegraphics[width=0.5\textwidth]{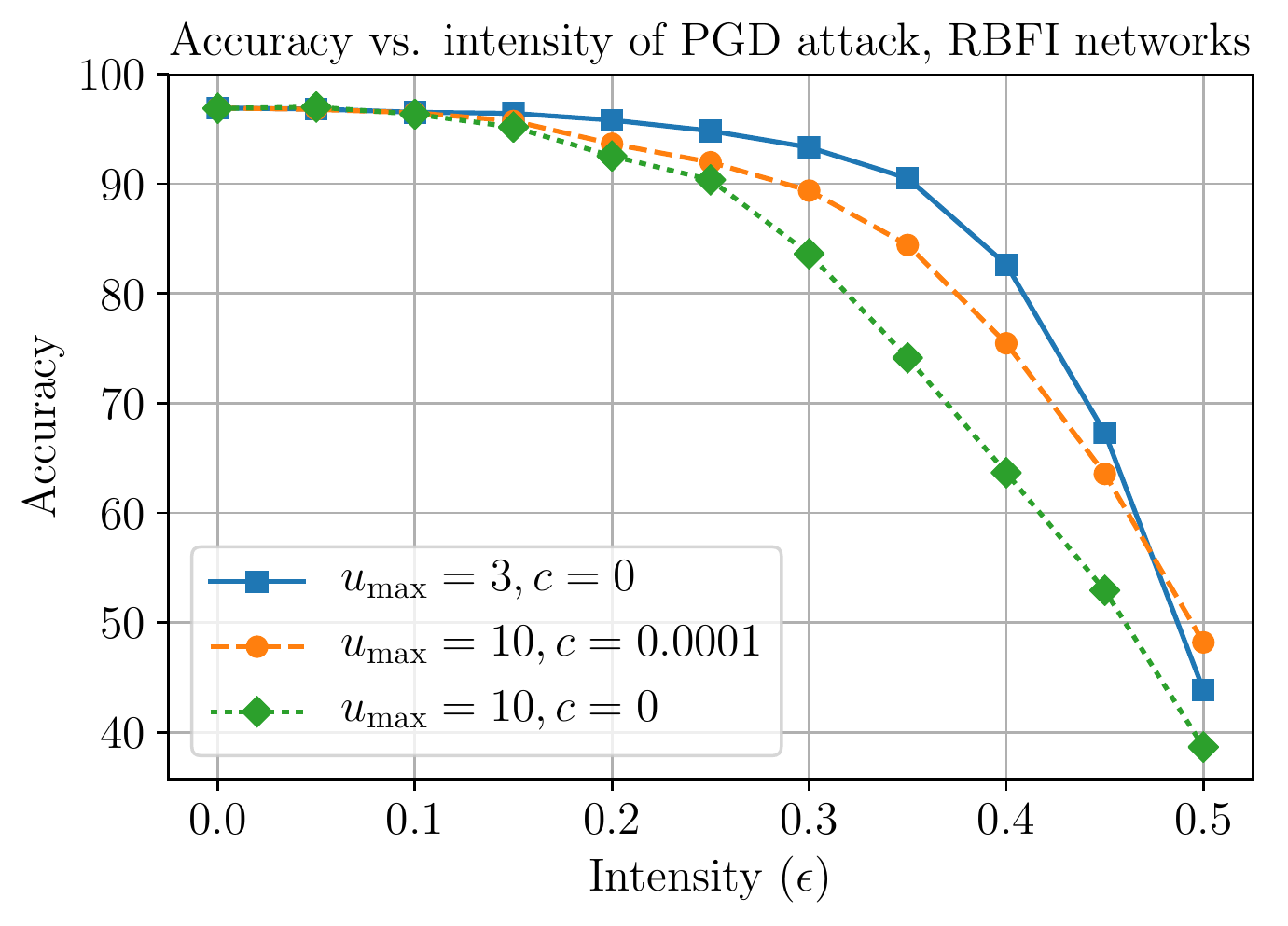}
\caption{Performance of $\rbfnet(512,512,512,10 \mid \und, \oder, \und, \oder)$ networks trained for 30 epochs, according to the $u_{\max}$ bound used for the $\vecu$ vectors, and according to the amount $c$ of regularization used.}
\label{fig-3-vs-10}
\end{figure}

We experimented with using $c \hat{s}$ as regularization loss, for a constant $c \geq 0$ and $\hat{s}$ as in (\ref{eq-sensitivity-network-rbf}). 
In Figure~\ref{fig-3-vs-10} we compare the robustness to PGD attacks of \rbfinf\ networks with $u$-bound 3, with that of \rbfinf\ networks with $u$-bound 10 trained without regularization and with $c = 0.0001$. 
We see that a small amount of regularization helps networks with higher $u$-bound to recover most of their robustness to attacks. 
We used PGD attacks, rather than FGSM or I-FGSM attacks, in the comparison, as \rbfinf\ networks are generally impervious to the latter. 
The regularization we used is fairly simple; it is likely that more sophisticated regularization would yield better results.

In Table~\ref{table-sensitivity} we report the sensitivity values computed via (\ref{eq-sensitivity-network-relu}) and (\ref{eq-sensitivity-network-rbf}) for different networks.  
As we see, the bounds do not seem to be very useful in comparing the resistance to attacks of neural networks of different architectures. 
ReLU networks trained with regularization can have sensitivities in the few hundreds, comparable to those of \rbfinf\ networks, and yet they are very susceptible even to the simple FGSM attack, as shown in Figure~\ref{fig-regularization}.

\begin{table}
\centering
\begin{tabular}{l|r|r|rl}
Network & $u_{\max}$ & $c$ & \multicolumn{2}{c}{Sensitivity bound} \\ \hline \hline
$\rbfnet(512,512,512,10 \mid \und,\oder,\und,\oder)$ & 3 & 0 & $43.72$ & $\pm 0.10$ \\
$\rbfnet(512,512,512,10 \mid \und,\oder,\und,\oder)$ & 10 & 0 & $1,072.90$ & $\pm 88.94$ \\
$\rbfnet(512,512,512,10 \mid \und,\oder,\und,\oder)$ & 10 & $0.0001$ & $231.90$ & $\pm 28.01$ \\ \hline 
ReLU(512,512,512,10) & & 0 & $275,296.24$ & $\pm 7,321.01$ \\
ReLU(512,512,512,10) & & $0.0003$ & $214.88$ & $\pm 3.83$ \\
ReLU(512,512,512,10) & & $0.0001$ & $383.30$ & $\pm 7.83$ \\
ReLU(512,512,512,10) & & $0.00003$ & $645.82$ & $\pm 13.22$ \\
ReLU(512,512,512,10) & & $0.00001$ & $985.57$ & $\pm 11.41$ \\ \hline
Sigmoid(512,512,512,10) & & 0 & $1,872,558.46$ & $\pm 27,969.37$
\end{tabular}
\caption{Sensitivity bounds, computed via (\ref{eq-sensitivity-network-relu}) and (\ref{eq-sensitivity-network-rbf}) for different networks and training regimes.}
\label{table-sensitivity}
\end{table}

\subsection{Deep \rbfinf\ Networks}
\label{subsec-deep}

Deeply non-linear models can be difficult to train in deep models consisting of many layers. 
To test the trainability of deep \rbfinf\ networks, we have experimented with training 128-128-128-128-128-10 networks for 20 epochs; for \rbfinf\ we used topology $\rbfnet(128,128,128,128,128,10 \mid \oder, \und, \oder, \und, \oder, \und)$.
ReLU and \rbfinf\ networks were easy to train: the former had $98.11 \pm 0.08\%$ accuracy, and the latter $95.53 \pm 0.13\%$ accuracy. 
At least with the methods we used (square-error loss and AdaDelta), which were the same for Sigmoid and \rbfinf\ networks, we were unable to train the Sigmoid network: the final accuracy was $11.24 \pm 0.34\%$, which is barely better than random.

\subsection{\rbfinf\ vs. RBF Units}

We have devoted the bulk of our attention to networks using \rbfinf\ units, rather than the more common RBF units defined by taking $\gamma=2$ in (\ref{eq-rbf}).
There are two reasons for this.
The first is theoretical. 
The sensitivity of a single RBF unit is given by $s = \sqrt{2/e} \cdot \norm{2}{\vecu}$, and thus grows with the square-root of the number of unit inputs; this is better than the linear growth with ReLU or sigmoid units, but not as good as the essentially constant behavior of \rbfinf\ units. 
The second reason behind our preference for \rbfinf\ units lies in the fact that RBF units turned out to be harder to train than \rbfinf\ units, even when using our pseudogradient techniques; this was particularly evident when we tried to train deep networks.  
This surprised us, as we thought that the more regular nature of the norm-2 metric would have helped training, compared to the infinity norm.

%% file: conclusions.tex
\section{Conclusions}

In this paper, we have shown that non-linear structures such as RBFI can be efficiently trained using artificial, ``pseudo'' gradients, and can attain both high accuracy and high resistance to adversarial attacks. 

Much work remains to be done. 
One obvious and necessary study is to build convolutional networks out of RBFI neurons, and measure their performance and resistance to adversarial attacks in image applications. 
Further, many powerful techniques are known for training traditional neural networks, such as dropout \cite{SrivastavaDropoutsimpleway2014}. 
It is likely that the performance of RBFI networks can also be increased by devising appropriate training regimes. 
Lastly, RBFIs are just one of many conceivable highly nonlinear architectures. 
We experimented with several architectures, and our experience led us to RBFIs, but it is likely that other structures perform as well, or even better. 
Exploring the design space of trainable nonlinear structures is clearly an interesting endeavor.

%% file: main.bbl
\newcommand{\etalchar}[1]{$^{#1}$}